%% file: main.tex
\def\arxiv{} 

\documentclass{article}

\usepackage{arxiv}

\usepackage[utf8]{inputenc} 
\usepackage[T1]{fontenc}    
\usepackage{hyperref}       
\usepackage{url}            
\usepackage{booktabs}       
\usepackage{amsfonts}       
\usepackage{nicefrac}       
\usepackage{microtype}      
\usepackage{lipsum}

\usepackage{xcolor}

\DeclareUnicodeCharacter{2212}{-}
\usepackage{caption}
\usepackage{subcaption}
\usepackage{textcomp}
\usepackage[
  separate-uncertainty = true,
  multi-part-units = repeat
]{siunitx}
\usepackage{longtable}
\usepackage[colorinlistoftodos]{todonotes}
\usepackage{pmboxdraw}
\usepackage{hyperref}
\usepackage{amsmath}
\usepackage[utf8]{inputenc}
\usepackage{spverbatim}
\usepackage{supertabular}
\usepackage{multicol}
\usepackage{booktabs}
\usepackage{comment}

\title{xDEEP-MSI: Explainable bias-rejecting microsatellite instability deep learning system in colorectal cancer}

\author{
  \href{https://orcid.org/0000-0001-6527-3059}{\includegraphics[scale=0.06]{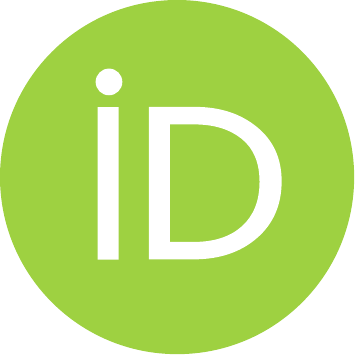}\hspace{1mm}Aurelia Bustos} \\
	AI Cancer Research Unit\\
	MedBravo\thanks{MedBravo \url{www.medbravo.org}}\\
	\texttt{aurelia@medbravo.org} \\
	\And
	Artemio Payá\\
	Pathology Department\\
	HGUA\thanks{Alicante University General Hospital, Alicante, Spain (HGUA)}, ISABIAL\thanks{Alicante Institute for Health and Biomedical Research (ISABIAL), Alicante, Spain} \\
	\And
	Andres Torrubia \\
	AI Cancer Research Unit\\
	MedBravo*\\
	\And
	Rodrigo Jover\\
	Gastroenterology Department\\
	HGUA$^\dagger$, ISABIAL$^\ddagger$\\
	\And
	Xavier Llor\\
	Department of Medicine\\
	Yale University\thanks{Cancer Center, Yale University, New Haven, Connecticut}\\
	\And
	Xavier Bessa\\
	Gastroenterology Department\\
	IMIM\thanks{Hospital del Mar, Hospital del Mar Medical Research Institute (IMIM), Barcelona, Spain}\\
	\And
	Antoni Castells\\
	Gastroenterology Department\\
	Hospital Clínic\thanks{Hospital Clínic, University of Barcelona, IDIBAPS, CIBERehd, Barcelona, Spain}\\
	\And
  \href{https://orcid.org/0000-0002-0560-1759}{\includegraphics[scale=0.06]{orcid.pdf}\hspace{1mm}Cristina Alenda} \\
	Pathology Department\\
	HGUA$^\dagger$, ISABIAL$^\ddagger$\\
	\texttt{alenda\_cri@gva.es} \\
   
}

\begin{document}

\maketitle

\begin{abstract}
We present a system for the prediction of microsatellite instability (MSI) from H\&E images of colorectal cancer using deep learning (DL) techniques customized for tissue microarrays (TMAs). The system incorporates an end-to-end image pre-processing module that produces tiles at multiple magnifications in the regions of interest as guided by a tissue classifier module, and a multiple-bias rejecting module. The training and validation TMA samples were obtained from the EPICOLON project and further enriched with samples from a single institution.
A systematic study of biases at tile level identified three protected (bias) variables  associated with the learned representations of a baseline model: the project of origin of samples, the patient's spot and the TMA glass where each spot was placed. A multiple bias rejecting technique based on adversarial training is implemented at the DL architecture so to directly avoid learning the batch effects of those variables. The learned features from the bias-ablated model have maximum discriminative power with respect to the task and minimal statistical mean dependence with the biases. 
The impact of different magnifications, types of tissues and the model performance at tile vs patient level is analyzed. The AUC at tile level, and including all three selected tissues (tumor epithelium, mucine and lymphocytic regions) and 4 magnifications, was 0.87±0.03 and increased to 0.9±0.03 at patient level. To the best of our knowledge, this is the first work that incorporates a multiple bias ablation technique at the DL architecture in digital pathology, and the first using TMAs for the MSI prediction task.
\end{abstract}


\textbf{\textit{Non-standard abbreviations and acronyms}} $BE$, batch effect module learners; $FE$, feature extractor backbone; $MSI$, Microsatellite Instability module learner; $dc$, squared distance correlation.

\graphicspath{{figs/}}

\section{Introduction}
\label{sec:intro}
Approximately 3\% of colorectal cancers (CRC) arise in the context of Lynch syndrome (LS), where the patient has a germline mutation in a DNA mismatch repair(MMR) gene \citep{moreira2012identification}.Historically, CRC patients were tested for LS if they were at high risk according to clinical criteria, e.g. aged under 50 years or with a strong family history. Several clinicopathologic criteria (e.g. Amsterdam criteria, revised Bethesda guidelines) were used to identify individuals at risk for Lynch syndrome or eligible for tumor-based MSI testing\citep{pinol2005accuracy}. However, a large proportion of LS patients were missed by this strategy \citep{west2021lynch}. Currently, new diagnostics guidance are recommending that all patients with newly diagnosed CRC be screened for LS. Universal tumor-based genetic screening for Lynch syndrome, with MSI or IHC testing of all CRCs regardless of age, has greater sensitivity for identification of Lynch syndrome as compared with other strategies \citep{moreira2012identification}. The pathway includes testing tumour tissue for defective MMR(dMMR) by either microsatellite instability (MSI) testing or immunohistochemistry (IHC) for the MMR proteins MLH1, PMS2, MSH2 and MSH6. Tumours showing MSI or MLH1 loss should subsequently undergo BRAF mutation testing followed by MLH1 promoter methylation analysis in the absence of a BRAF mutation. Patients with tumours showing MSH2, MSH6 or isolated PMS2 loss, or MLH1 loss/MSI with no evidence of BRAF mutation/MLH1 promoter hypermethylation, are referred for germline testing if clinically appropriate.

MSI is not specific for LS, and approximately 15 percent of all sporadic CRCs and 5 to 10 percent of metastatic CRCs demonstrate MSI due to hypermethylation of MLH1 \citep{nunes2020molecular, jenkins2007pathology,west2021lynch}. Sporadic MSI-H CRCs typically develop through a methylation pathway called CpG island methylator phenotype (CIMP), which is characterized by aberrant patterns of DNA methylation and frequently by mutations in the BRAF gene. These cancers develop somatic promoter methylation of MLH1, leading to loss of MLH1 function and resultant MSI. The prevalence of loss of MLH1 expression in CRC increases markedly with aging and this trend is particularly evident in women \citep{kakar2003frequency}.

While guidelines set forth by multiple professional societies recommend universal testing for dMMR/MSI \citep{sepulveda2017molecular}, these methods require additional resources and are not available at all medical facilities, so many CRC patients are not currently tested \citep{shaikh2018mismatch}. 

Since the last two decades, certain histology-based prediction models that rely on hand-crafted clinico-pathologic feature extraction - such as age <50, female sex, right sided location, size >= 60 mm, BRAF mutation, tumor infiltrating lymphocytes (TILs), a peritumoral lymphocytic reaction, mucinous morphology and increased stromal plasma cells - have reported encouraging performance but has not been sufficient to supersede universal testing for MSI/dMMR \citep{hildebrand2021artificial}. 
Measurement of the variables for MSI prediction, requires significant effort and expertise by pathologists, and inter-rater differences may affect the perceived reliability of histology-based scoring systems \citep{hildebrand2021artificial}. However, this work is fundamental to the premise that MSI can be predicted from histology, which was recently proposed as a task for deep learning from digital pathology \cite{zhang2018adversarial, kather2019deep}.

Research on deep learning methods to predict MSI directly from hematoxylin and eosin (H\&E) stained slides of CRC have proliferated in the last two years, \citep{zhang2018adversarial, kather2019deep,kather2020pan, cao2020development, yamashita2021deep,echle2020,9474704,pericles} with reported accuracy rapidly improved on most recent works \citep{yamashita2021deep,echle2020}.

If successful, this approach could have significant benefits, including reducing cost and resource-utilization and increasing the proportion of CRC patients that are tested for MSI. Additional potential benefits are to increase the capability of detecting MSI over current methods alone. Some tumors are either dMMR or MSI-H but not the other. Testing of tumors with only immunohistochemistry (IHC) or polymerase chain reaction (PCR) will falsely exclude some patients from immunotherapy \cite{SAEED202110}. A system trained on both techniques could overcome this limitation, obviating co-testing with both MMR IHC and MSI PCR as an screening strategy for evaluating the eligibility status for immunotherapy. 

Among current limitations, the developed systems so far are not able to distinguish between somatic and germline etiology of MSI, such that confirmatory testing is required. Another important limitation is generalizability due to batch effects, while those systems have proved excellent performance on well curated cohorts that are similar to training data, the performance is not robust to differing patient and tissue characteristics. This limitation is evident from the performance deterioration when systems trained on a single datasource are tested on external datasets \cite{yamashita2021deep}.  This limitation can be palliated by using larger multi-institutional datasets from different institutions for training, as shown by \cite{echle2020clinical} that estimated 5000 to be the optimal number of patients needed for this specific task. Nonetheless, compiling such international large scale datasets is costly and unfeasible in most cases and does not eliminate batch effects. Regardless of the dataset size and number of contributing sites, the propensity for over-fitting of digital histology models to site level characteristics is incompletely characterized and is infrequently accounted for in internal validation of deep learning models \citep{howard2020impact}. For example assessments of stain normalization and augmentation techniques have focused on the performance of models in validation sets, rather than true elimination of batch effect \citep{anghel2019high,echle2020}. In addition to staining techniques, batch effects originate from other multiple reasons, such as digitization of slides, variations due to scanner calibration and choice of resolution and magnification. Batch effects in training, validation and testing, must be accounted for to ensure equitable application of DL. Batch effects leads to overoptimistic estimates of model performance and methods to not only palliate but to directly abrogate this bias are needed.\citep{howard2020impact}

Herein, we present a novel approach in digital pathology to eliminate the batch effects at the deep learning architecture following the methodology described in \citep{zhao2020training}, where by means of an adversarial training and bias distillation regime, the model avoid learning undesirable characteristics of datasets such as the project or other protected variables. We extend this methodology further by first, systematically assessing and quantifying the spurious associations of protected variables (biases) on the network and second, leveraging a multiple bias-ablation architecture in the model.

The remainder of the paper is as follows. Section \ref{sec:methods} describes the methodology employed, including the study population, the image pre-processing module, partitioning methods and deep learning architecture, the identification of biases, the bias-ablation system and training regime. Section \ref{sec:results} first describes the results of the tissue classifier module, the image dataset obtained after pre-processing, the results of the bias identification, the results of the MSI-status classifier module at image level with the demonstration of batch effect distillation, performance results at patient level, analysing the impact of types of tissues included and image magnifications, and the explainability of predictions. Finally, Section \ref{sec:discussion} addresses the discussion, conclusions and future work.

\section{Material and methods}

\label{sec:methods}

\subsection{Study Population, Data Collection and Ground Truth Ascertainment}
The study population initially consisted of the H\&E of 57 tissue-microarrays (TMAs) that included cylindrical tissue samples of 1 mm diameter each (spots), in duplicate from all patients prospectively collected in the EPICOLON project. In this technique, each TMA consists of multiple spots placed in one microscope glass slide. This population-based, observational, cohort study included 1705 patients with CRC from 2 Spanish nationwide multi-center studies: EPICOLON I and EPICOLON II. EPICOLON I included consecutive patients with a new diagnosis of CRC between November 2000 and October 2001 with the main goal of estimating the incidence of LS in Spain\citep{pinol2004frequency}. EPICOLON II also included consecutive patients with newly diagnosed CRC between March 2006 and December 2007 and from 2009 to 2013 with the aim of investigating different aspects related to the diagnosis of hereditary CRC \citep{abuli2010susceptibility}. Both studies were approved by the institutional review boards of the participating hospitals. The overall MSI frequency in the EPICOLON project was 7.4\%.  

\begin{figure}
    \centering
    \includegraphics[width=0.80\textwidth]{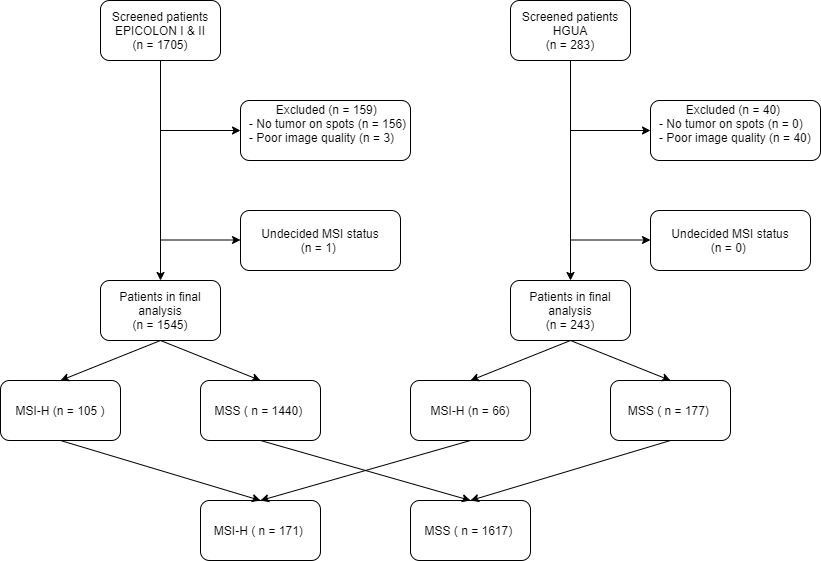}
    \caption{Study sample flowchart}
    \label{fig:Sample}
\end{figure}

The population was further expanded with 283 additional patients retrospectively obtained from the \textit{Hospital Universitario de Alicante, Spain} (HGUA) from 2017 onwards, which -once pre-processed- added 66 MSI-H and 177 MSS cases to the final study population as shown in Fig. \ref{fig:Sample}. The corresponding H\&E images were provided in 15 TMAs that included 1 mm spots in duplicate for each patient. 

For the task of MSI prediction each patient and corresponding spots were labeled as MSI-H vs MSS. MSI-H was defined as tumour-tissue testing defective MMR(dMMR) by either microsatellite instability (MSI) testing or immunohistochemistry (IHC) for the MMR proteins MLH1, PMS2, MSH2 or MSH6. Tumours showing MSI or MLH1 loss underwent BRAF mutation testing followed by MLH1 promoter methylation analysis in the absence of a BRAF mutation. Briefly, patients with tumours showing MSH2, MSH6 or isolated PMS2 loss, or MLH1 loss/MSI with no evidence of BRAF mutation (regardless of presence or not of MLH1 promoter hypermethylation), were labeled as MSI-H and all others as MSS. 

TMAs were scanned with 
at magnification x40 corresponding to a maximal resolution of 0.25 microns per pixel (MPP) 

This project was approved by the institutional research committee CEIm PI2019-029 from ISABIAL and both the images and associated clinical information was previously anonymized. The data from EPICOLON project and HGUA are not publicly available, in accordance with the research group and institutional requirements governing human subject privacy protection.

The NCT-CRC-HE-100K and CRC-VAL-HE-7K datasets, consisting of 100,000 non-overlapping image patches from hematoxylin \& eosin (H\&E) stained histological images of human colorectal cancer (CRC) and normal tissue labeled with 9 tissue classes - adipose (ADI), background without tissue (BACK), debris (DEB), lymphocytes (LYM), mucus (MUC), smooth muscle (MUS), normal colon epithelium (NORM), cancer-associated stroma (STR) and colorectal adenocarcinoma epithelium (TUM) -  were used to train a model in the task of tissue class prediction. All images were 224x224 pixels (px) at 0.5 MPPs. All images were color-normalized using Mazenko's method \citep{macenko2009method}. All image tiles for the NCT-CRC-HE-100K and CRC-VAL-HE-7K datasets are available online \footnote{ \url{https://zenodo.org/record/1214456\#.XcNpCpNKjyw}}.  



\subsection{Preprocessing module}
The image pre-processing module consisted of a TMA-customized dynamic extractor of tiles of 400x400 pixels corresponding to adjacent regions without overlap parameterized at different magnifications (x40, x20, x10, x5, x0) linked to a DL tissue classifier (described in see Section \ref{sec:DL}) that filtered spots without viable tissue and at the same time selected the regions of interest based on type of tissues.

\subsection{Deep Learning architecture}
\label{sec:DL}
The DL architecture (see Fig. \ref{fig:Architecture}) consisted of an end-to-end deep learning system linking three types of modules: 1) A classifier of 9 types of tissues at tile level: adipose (ADI), background without tissue (BACK), debris (DEB), lymphocytes (LYM), mucus (MUC), smooth muscle (MUS), normal colon epithelium (NORM), cancer-associated stroma (STR) and colorectal adenocarcinoma epithelium (TUM) 2) the MSI status classifier $MSI$  and 3) the batch effect module learners $BE$, which has as many learners or heads as required to account for multiple biases. 

The classifier of tissues, had as a convolutional backbone a Resnet34 pre-trained on Imagenet, and two hidden layer of dimension 512 with ReLU as the activation function, and 9 out features. This module was trained with the NCT-CRC-HE-100K and CRC-VAL-HE-7K datasets on the final task of learning the 9 types of tissue. Once trained, it was used both to select the regions of interest (ROIs) based on type of tissues in the pre-processing module and as the feature extractor backbone $FE$ in the end-to-end system (removing the head) connected with the $MSI$ and $BE$ modules. The $FE$ yielded 512 intermediate features.

Both the $MSI$ and all $BE$ modules have two hidden layers of dimension 512 with ReLU as the activation function, and 2 out features.

The DL system without the $BE$ modules is referred herein as the baseline model, and when including the $BE$ modules, is referred as the bias-ablated model.

All code was programmed using the Python programming language. 
Machine and deep learning methods were implemented using FastAI \citep{howard2018fastai} and PyTorch. QPath v0.2.3 \citep{bankhead2017qupath} was used for annotation purposes.

\subsection{Bias identification}
To systematically identify and quantify the effect of potential biases interfering with the MSI prediction task, the baseline model, which as described in Section \ref{sec:DL} included only the feature extractor backbone $FE$ connected to the MSI status classifier $MSI$ module, was trained on the study sample using 5-fold cross-validation for 3 epochs. The squared distance correlation $dc$ \citep{szekely2007measuring} was computed on each batch-iteration between the extracted features from the $FE$ and three putative biases (study project, patient, and glass) followed by backward selection to uncover hidden interactions between biases. The $dc$ is a measure of dependence between random vectors analogous to product-moment covariance and correlation, but unlike the classical definition of correlation, distance correlation is zero only if the random vectors are statistically independent. \citep{szekely2007measuring}

The three biases considered for testing were the study project, the patient and the TMA glass based on the following reasoning:
First, since MSI-H patients were significantly over-represented in the subset of cases obtained from a single hospital (34\% in HGUA vs 7.4\% in EPICOLON), and also samples were obtained in different years on each project (2017 onwards in HGUA vs prior to 2013 in EPICOLON) in this study the contributing project becomes a potential task bias; i.e.  prediction of MSI status may be dependent on project instead of on the image bio-markers of MSI status.
Second, tiles from the same patient commonly share observable visual patterns originated from both tissue characteristics, slicing direction and staining differences in laboratory procedures. As the model is designed to be trained at tile level, even if no patient is at the same time in the training and validation set, the model would learn those patterns as shortcuts for the MSI task, hence being a potential reason of model over-fitting. Over-fitting was indeed verified when the baseline model was trained more than 3 epochs observing an ever decreasing training loss near zero with no improvements in the validation loss.
Third, each TMA agglutinates sample spots from tens of patients which are placed between an underlying and a cover glass. The unique characteristics associated with the digitization of each TMA-glass may be statistically associated with MSI status if the distribution of classes across TMAs is not uniform. 

\subsection{Bias ablation }
\begin{figure*}
\centering
     \begin{subfigure}[b]{0.45\textwidth}
         \centering
         \includegraphics[width=\textwidth]{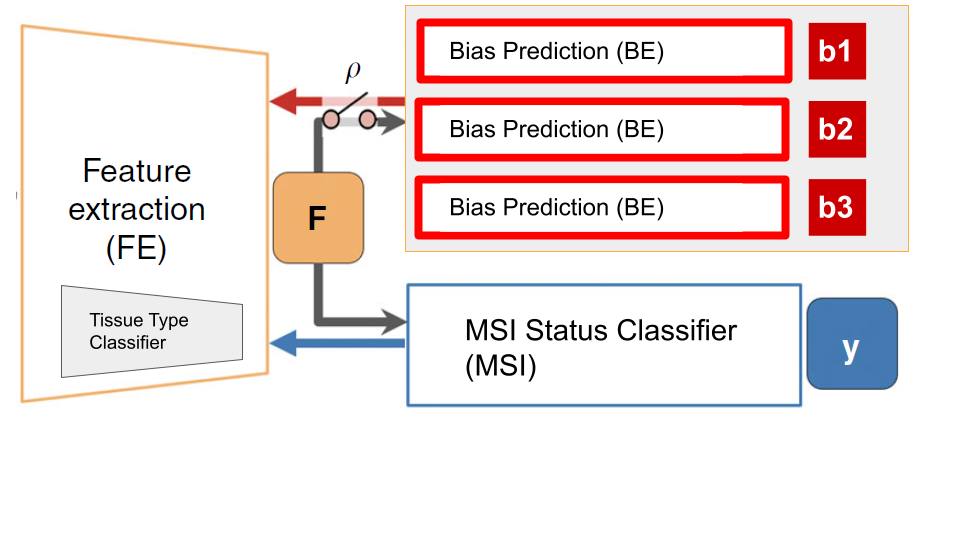}
         
     \end{subfigure}
     \begin{subfigure}[b]{0.45\textwidth}
         \centering
         \includegraphics[width=\textwidth]{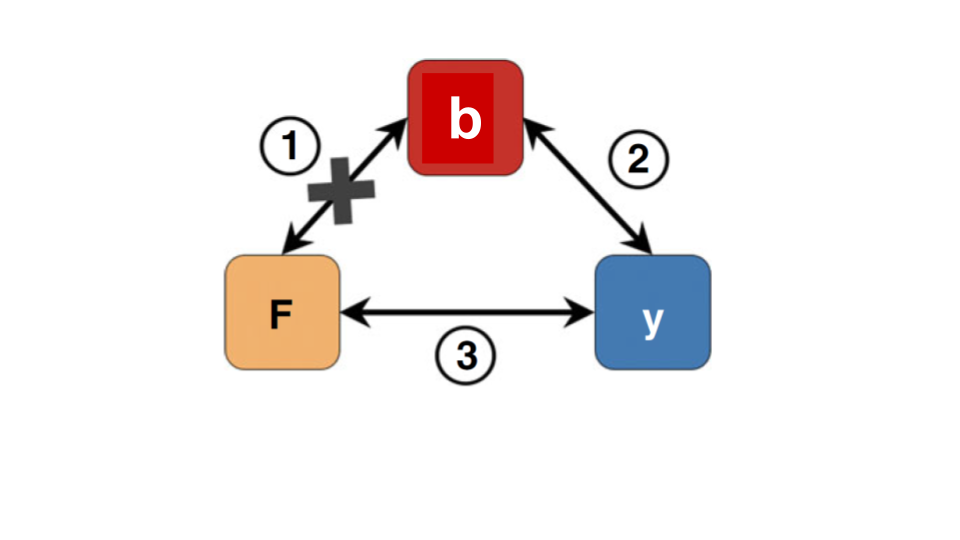}
         
     \end{subfigure}
     \hfill
    \caption{Network architecture:  a) The deep neural network architecture is composed of three modules, FE learns features, $F$, that successfully classify the input in the outcome $y$ using $MSI$ while being invariant (statistically independent and conditioned by $\rho$) to the biases variables, $b_n$, using the adversarial components BE and the adversarial loss.  b) The bias variables $b_n$, responsible for multiple batch effects, influence both the output $y$ (i.e., \raisebox{.5pt}{\textcircled{\raisebox{-.9pt} {2}}}, MSI status classification) and the input $X$, from which feature $F$ is extracted (i.e., \raisebox{.5pt}{\textcircled{\raisebox{-.9pt} {1}}}). The MSI classifier deems to find the relation \raisebox{.5pt}{\textcircled{\raisebox{-.9pt} {3}}} to enable prediction of the output labels while the adversarial components aim to remove the direct dependency between $F$ and $b_n$. Figure adapted from \citep{zhao2020training}
  }
    \label{fig:Architecture}
\end{figure*}

Once biases were identified and quantitatively characterized, the ablation of each bias $b_n$, was implemented through an adversary training and distillation bias regime following the approach described in \cite{zhao2020training} (see Fig. \ref{fig:Architecture}), but differently from this work, the ablation was not limited to only one but to all the biases identified. 

Namely, given the input image $X$, the $FE$ module extracts a feature vector $F$, on top of which the $MSI$ module predicts the class label $y$. To guarantee that $F$ is not biased to the multiple $b_n$, each corresponding $BE_n$ module back-propagate, in a consecutive way, the loss to $FE$ adversarially, i.e. as $-\lambda L_{be_n}$. It results in features that minimize the classification loss of the $MSI$ module while maintaining the least statistical dependence on each of the bias $b_n$.

Each of the $BE$ modules is trained on a $y$-conditioned cohort, i.e., samples of the training data whose $y$ values (MSI labels) are confined to a specific group (referred as $\rho$ in Fig. \ref{fig:Architecture}). Consequently, the features learned by the system are predictive of $y$ while being conditionally independent of the batch effect originated by each of the biases. On the implemented architecture, the system would learn to separate MSI-H vs MSS samples by training each $BE_n$ only on the MSS group to correctly model the batch effects on the samples. We perform the adversarial training of each of the $BE_n$ only on the MSS group.

During the end-to-end system training a min-max game, is defined between two networks. The classification loss $L_{msi}$ is defined by a cross-entropy: 
\begin{equation}
L_{msi}(X,y;\theta_{fe},\theta_{msi}) = - \sum_{i=1}^{N}\sum_{m=1}^{M}y_{im}log(\hat{y}_{im})
\end{equation}
where X and y are the input images and corresponding msi target labels, respectively, $N$ is the number of training pairs (X,y), $M$ is the number of classes to predict (two, MSI-H vs MSS) and $\hat{y}$ is the predicted MSI label.

Each batch effect or bias loss $L_{be_n}$ is based on the squared Pearson correlation $corr^2$: 
\begin{equation}
L_{be_n}(X,y;\theta_{fe},\theta_{be_n}) = - \sum_{k=1}^{K}corr^2(b_k, \hat{b}_{k})
\end{equation} 
where $b_k$ defines the vector of the bias across all N training inputs. The statistical dependence is removed by pursuing a zero correlation between $b_k$ and $\hat{b}_k$ through adversarial training. 

The overall objective of the end-to-end network is defined as:

\begin{equation}
\underset{\theta_{fe},\theta_{msi}}{min}\underset{\theta_{be}}{max} L_{msi}(X,y;\theta_{fe},\theta_{msi}) - \lambda\sum_{b=1}^{B} L_{be_n}(X,b_n;\theta_{fe},\theta_{be_n}) 
\end{equation}
where B is the number of protected variables or biases $b$

Specifically, in each iteration, first we back-propagate loss $L_{msi}$ to update $\theta_{fe}$ and $\theta_{msi}$. Second, for each $b_n$, we fix $\theta_{fe}$ and then minimize the $L_{be_n}$ loss to update their corresponding $\theta_{be_n}$. Finally, we fix $\theta_{be_n}$ and then maximize the $L_{be_n}$ loss to update $\theta_{fe}$, hence distilling all the biases from $\theta_{fe}$. In
this study, each $L_{be_n}$ depends on the correlation operation, which
is a population-based operation, as opposed to individual level
error metrics such as cross-entropy or MSE losses.
Therefore, we calculate the correlations over each training
batch as a batch-level operation. In conclusion, $FE$ extracts features that minimize the classification criterion, while ‘fooling’ all $BE$  modules (i.e., making each $BE$ incapable of predicting
their corresponding bias ). Hence, the saddle point for this
objective is obtained when the parameters $\theta_{FE}$ minimize the
classification loss $L_{msi}$ while maximizing the $L_{be_n}$ loss of all the $BE_n$ modules.

\subsection{Partitions, model training and metrics}

For model training, the dataset was split 80/20 for training and validation applying 5-fold cross-validation and guaranteeing on each fold that the images of each patient only belonged to one set (either training or validation but not both).
To address imbalance in MSI status as well as in the number of tiles for each patient, we applied a composite weighted random sampling for both criteria, resulting in a balanced training set for both patient and MSI label simultaneously.
Tiles were resized to 224 x 224 pixels and color was normalized following Mazenko method \citep{macenko2009method}. In addition to Mazenko, experiments were done with and without an additional color normalization with statistics computed from the EPICOLON image dataset which included 25 different hospitals. 
Data augmentation at training time consisted in random rotations up to 90º, dihedral flips with probability of 0.5, a perspective warping of maximum 0.2, and hue variations of maximum 0.15. Of note, training sets included all magnifications so that the network could be trained simultaneously on higher tissue architecture patterns as well as on cellular-level features including nuclear characteristics. The batch size was 512 images. 

The statistical dependency between learned features and each of the selected biases was monitored during model training with the squared distance correlation $dc$. Principal component analysis (PCA) was used to assess how the spatial representations of the learned features were affected by the protected (bias) variables before and after bias ablation. Metrics used to assess the performance of the MSI classifier include AUC, balanced accuracy, sensitivity, specificity, positive predictive value, negative predictive value, false positive rate and false negative rates. Metrics dependent on MSI-H prevalence are calculated assuming 15\% in the real population.

\subsection{Explainable Methods}
SHAP (SHapley Additive exPlanations) \citep{lundberg2017unified} values were used to provide a means of visually interpreting the topology and morphology of features that influence predictions. The goal of SHAP is to explain the prediction of an instance by computing the contribution of each feature to the prediction. The SHAP explanation method computes Shapley values from coalitional game theory.

\section{Results}

\label{sec:results}
\subsection{Experiment 1: Tissue classifier}
\begin{figure}[h]
\centering
\begin{subfigure}{0.45\textwidth }
\includegraphics[width=0.9\linewidth, height=.16\textheight]{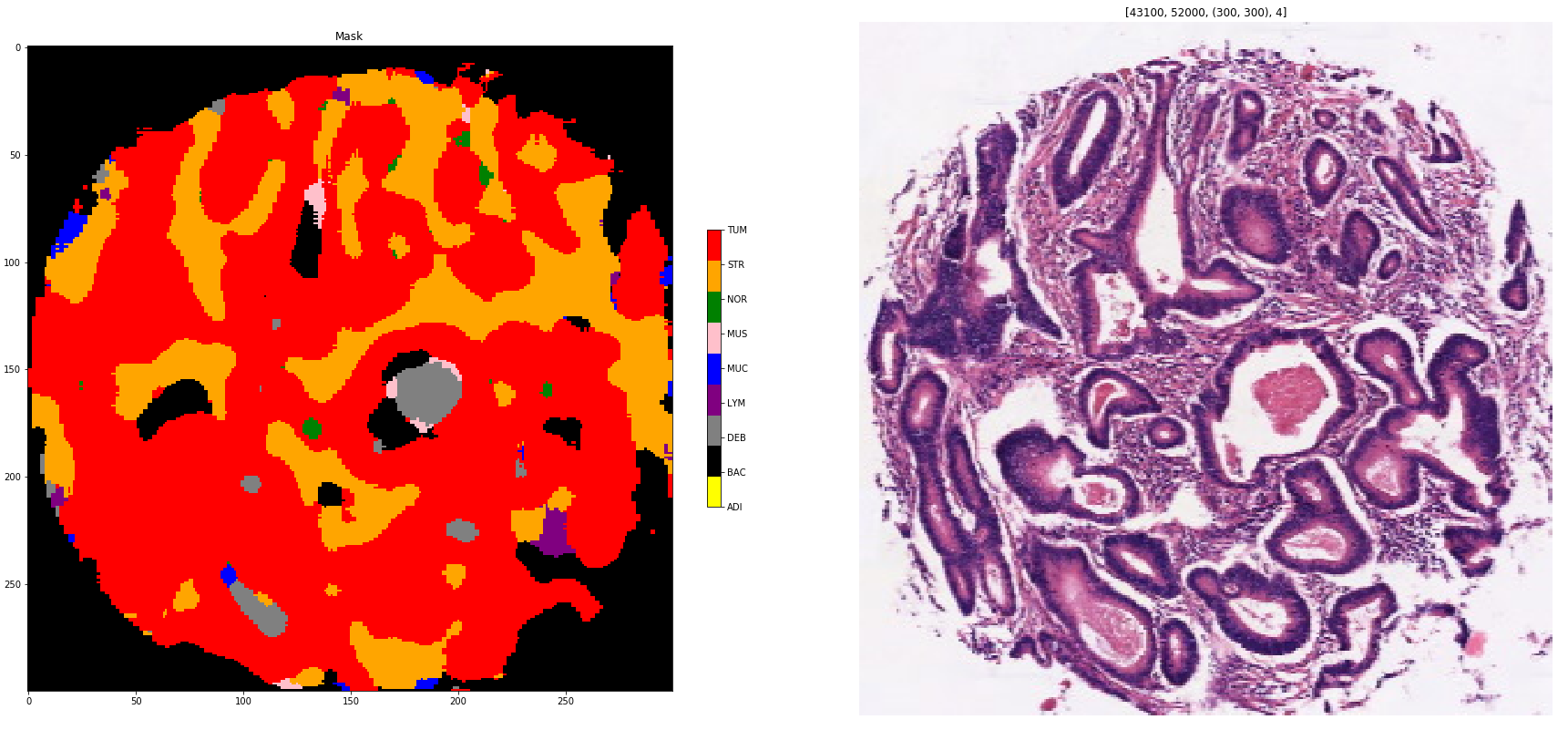} 
\caption{}
\label{fig:Spot1}
\end{subfigure}
\begin{subfigure}{0.45\textwidth}
\includegraphics[width=0.9\linewidth, height=.16\textheight]{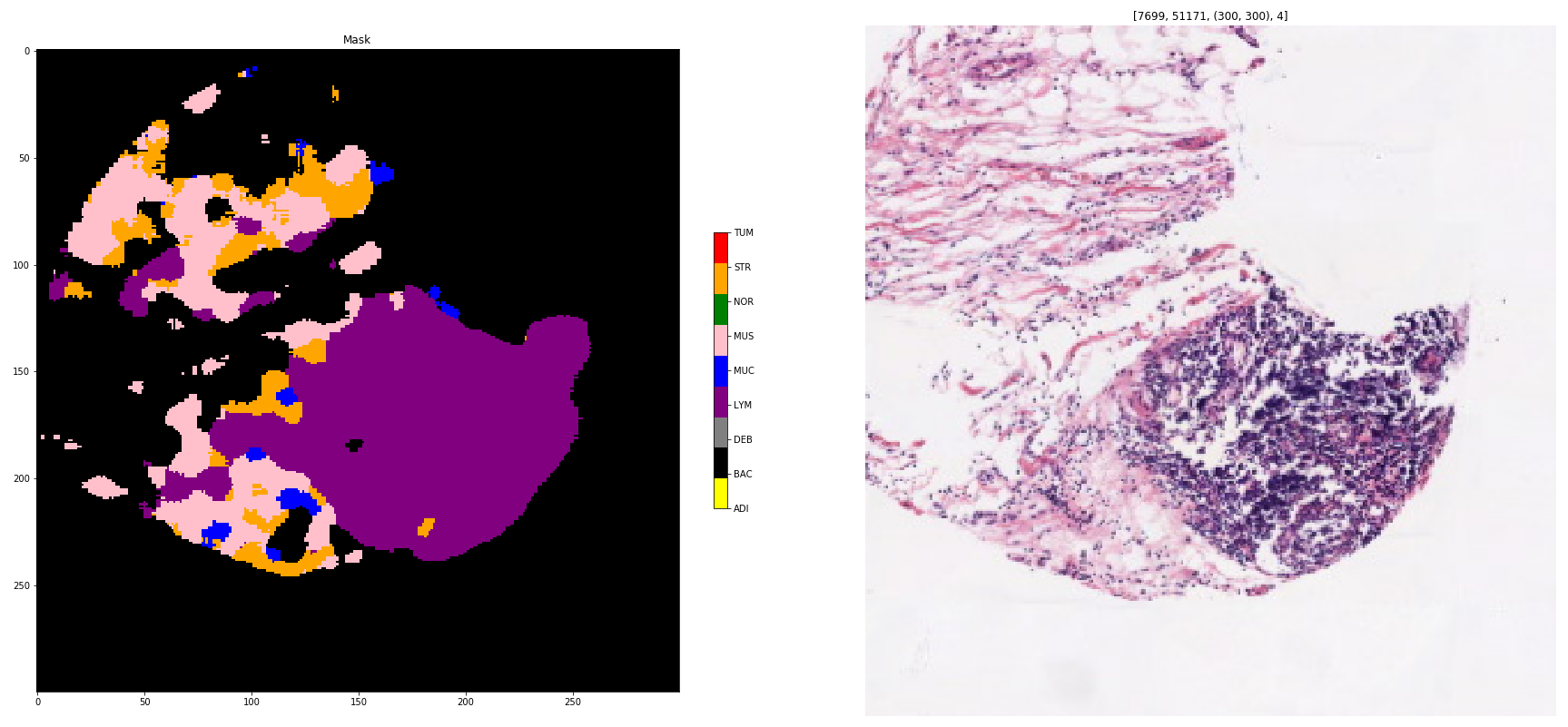}
\caption{}
\label{fig:Spot2}
\end{subfigure}
\begin{subfigure}{0.45\textwidth}
\includegraphics[width=0.9\linewidth, height=.16\textheight]{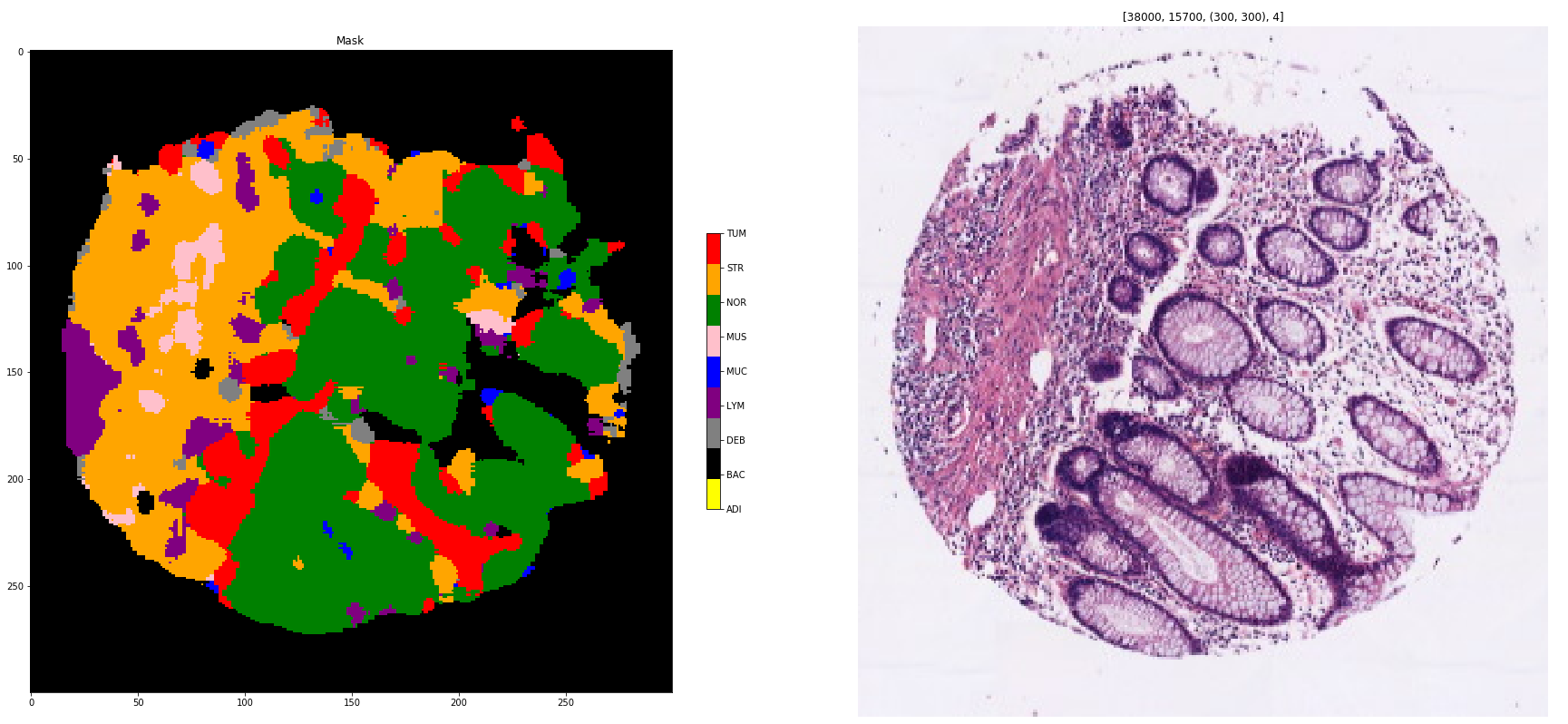} 
\caption{}
\label{fig:Spot3}
\end{subfigure}
\begin{subfigure}{0.45\textwidth}
\includegraphics[width=0.9\linewidth, height=.16\textheight]{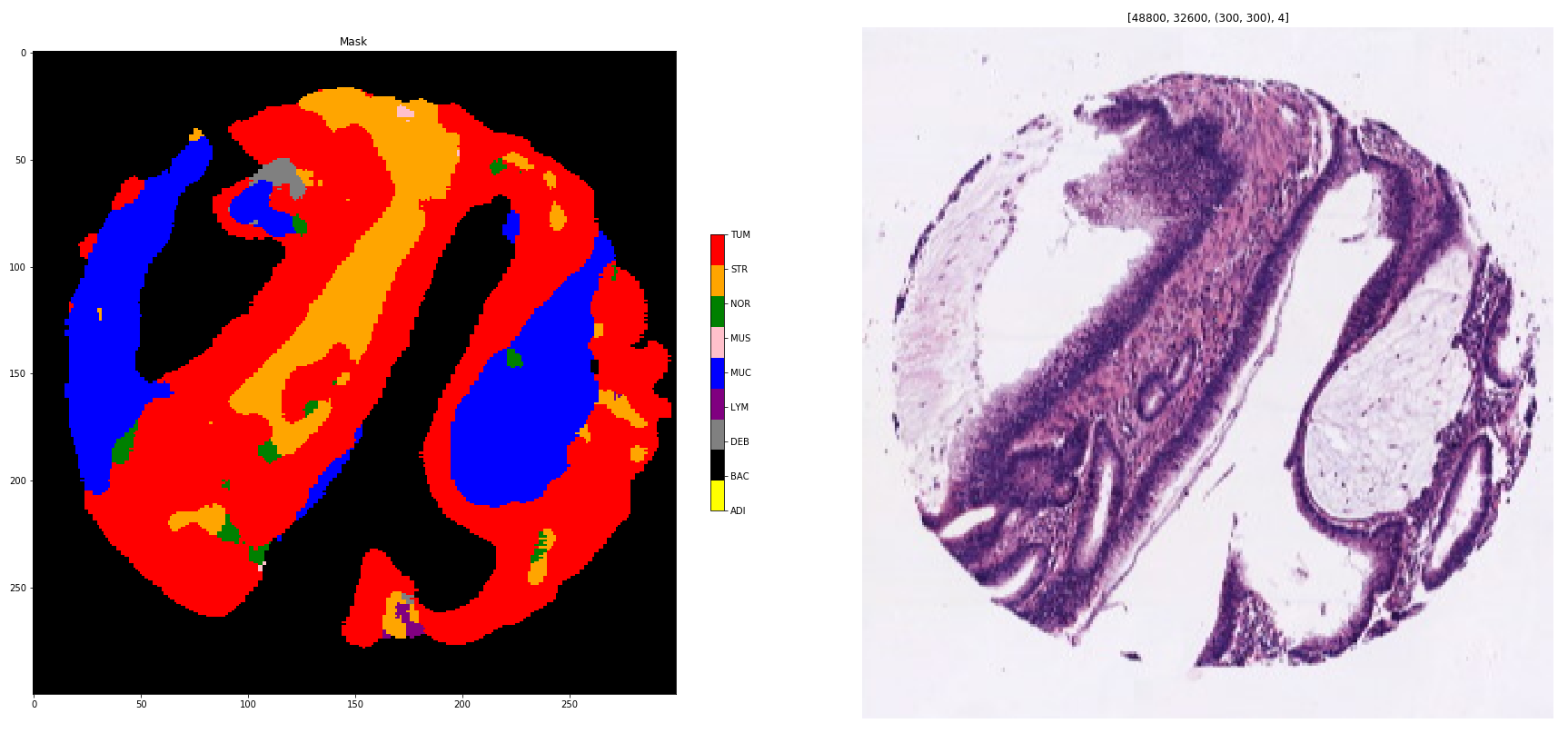}
\caption{}
\label{fig:Spot4}
\end{subfigure}
\caption{Examples of spots with each tissue region coded with different colors: tumor epithelium (red), stroma (orange), normal epithelium (green), mucine (blue), muscular fibers (pink), lymphocytic infiltrates (purple), debris (grey), adipose tissue (yellow) and background (black). Original samples are placed on the right and color masks on the left. The tissue class for each pixel is computed as the majority tissue class of partial overlapping tiles applying a sliding window to raster each spot. Tiles were input to the tissue classifier module at x20 magnification.}
\label{fig:Tissue_spots}

\end{figure}

The tissue classifier module reached an AUC of 0.98 in the validation set.
This module was capable of classifying the different regions of the image (tumor epithelium, stroma, normal epithelium, mucine, muscular fibers, lymphocytic infiltrates, debris, adipose tissue and background) as shown in Fig. \ref{fig:Tissue_spots}

After the automatic filtering done by the pre-processing module, the final study sample totaled 1788 patients (171 MSI-High). The image dataset consisted of 1,065,479 tiles or adjacent regions without overlap including all magnifications and all types of tissues. The tissue classifier module was then used in inference for the pre-selection of the regions of interest restricted to tumor epithelium, lymphocytic infiltrates and mucine totaling 523,624 tiles, linked to the
entrance to the $MSI$ module classifier. On this filtered dataset, the distribution of tiles by each magnification is shown in Tab. \ref{tab:dist_magnification}

\begin{table}[]
    \centering
    \caption{Distribution of tiles by magnification. Tiles were filtered by regions of interest restricted to tumor epithelium, lymphocytic infiltrates and mucine.}
    \begin{tabular}{c|c}
\textbf{Magnification} & \textbf{Number of tiles} \\  \toprule
x40 &   386524\\
x20 &   100569\\
x10 &    23630\\
x5  &    9954\\
x0  &    2947\\ \midrule
All(total) & 523624\\
\bottomrule

    \end{tabular}
    \centering
    
    \label{tab:dist_magnification}
\end{table}

\subsection{Experiment 2: Bias identification, interaction between variables and bias ablation implementation }

First a baseline model was trained for 10 epochs and over-fitting was observed with an ever decreasing training loss approaching zero with no improvements in the validation loss. The lack of generalizability in validation was suspected to the presence of biases described in Section \ref{sec:methods}.

Second, a baseline model was trained for 3 epochs, with early stopping, and the statistical dependence between the representations learned by the feature extractor backbone with regards to the target task (MSI classification) and each of the suspected bias was measured in the entire training cohort with the squared distance correlation $dc$. As shown in Tab. \ref{table:dc bias-identification}, the learned features by the baseline network had surprisingly a higher $dc$ with the project variable (0.17) when compared with the MSI label (0.08). Subsequently, as a subgroup analysis, $dc$ was computed for each project cohort separately. At this step a hidden interaction emerged between the HGUA project cohort and the TMA glass and patient variables respectively, where the $dc$ increased up to 0.29 in both cases. The unexpectedly strong association between the TMA glass and the learned representations only occurring in the HGUA project was further investigated and it obeyed to the MSI label distribution of the spots among TMAs. Specifically, we found that many TMAs in the HGUA project had spots from only one class (either MSS or MSI-H, but not both). Moreover, as the glass covered the spots it meant that not only the background but also the tissue regions carried the glass characteristic patterns. This observation was relevant implying that removing background regions -even if necessary- is not enough to remove this bias. In contrast, all of the EPICOLON TMAs had a varying representation of spots of both classes, though always strongly unbalanced towards the MSS class. 

Third, once confirmed the presence of multiple bias variables (i.e. project, patient, and TMA glass variables), three corresponding bias-rejecting learners were attached to the end-to-end network applying an adversarial consecutive training regime for each head as explained in Section \ref{sec:DL}.  
After implementing the bias-ablation technique and retraining, we recomputed the statistical dependence between the representations learned by the bias-ablation model with respect to the target and each bias. As shown in Tab. \ref{table:dc bias-identification}, associations of the learned features with the MSI status were strengthened, increasing from 0.08 to 0.21 in the overall study set. Conversely, the dependence with three biases was weakened, decreasing from 0.17 to 0.02 with respect to the project bias and from 0.08 to 0.02 in the case of the patient and glass bias. When narrowing down to the HGUA project, the bias ablation resulted in a higher association of the learned features with the MSI status, but failed to disentangle their association with the glass and patient bias which still remained high. As explained above, in the HGUA project some TMAs were composed only of one single target class of spots, hence, the observed persistent statistical dependence supports that the attempt to decoupling those TMA glasses from the MSI status is unfeasible and that the glass may provide the network with an unfair shortcut for MSI status prediction.

\begin{table}[]
    \centering
    \caption{Statistical dependence between the learned features wrt the target task (MSI) and each bias quantitatively measured by the squared distance correlation ($dc$). Hidden interactions between the different biases are explored by subgroups by re-calculating the $dc$  for each project. The learned features from a baseline model are compared against a bias-ablated model with regards to their statistical dependency to the target task and each bias. In bold are highlighted the lowest distance correlations achieved for each bias. The bias-ablation model maximizes the association of the learned features with respect to the MSI task and minimizes the statistical mean dependence with the biases. }
    \begin{tabular}{c | c | c | c| c| c} \toprule
    \textbf{Model} & \textbf{Subgroup} & \textbf{MSI $dc$} & \textbf{Project Bias $dc$} & \textbf{Patient Bias $dc$} & \textbf{Glass Bias $dc$}\\ \midrule
    Baseline model & All & 0.08  & 0.17 & 0.08 & 0.07 \\
    Baseline model & EPICOLON Project & 0.04  & -
    & 0.02 & 0.07  \\
    Baseline model & HGUA Project & 0.46  & - & 0.29 &  0.29  \\ \midrule
    Bias-ablation model & All & 0.21  & \textbf{0.02} & 0.02 & \textbf{0.02} \\
    Bias-ablation model & EPICOLON Project & 0.15  & -
    & \textbf{0.01} & 0.07  \\
    Bias-ablation model & HGUA Project & 0.62  & - & 0.35 &  0.33  \\
    \bottomrule
\end{tabular}

\label{table:dc bias-identification}
\end{table}

\subsection{Experiment 3: Bias ablation analysis} 
\begin{figure}
    \centering
    \includegraphics[width=0.80\textwidth]{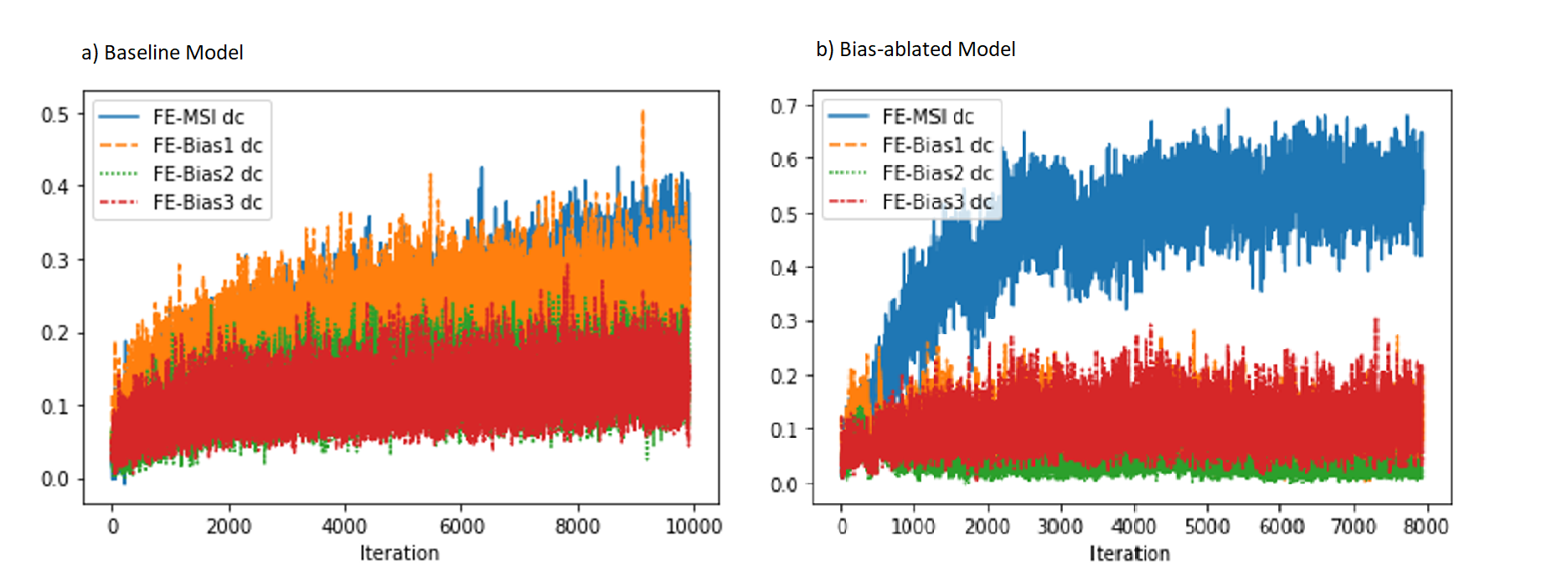}
    \caption{Statistical dependency, measured with the squared distance correlation ($dc$), between the learned features, MSI learning task and each bias (project, patient and glass bias) during the training process. After bias-ablation, the statistical dependency of the learned features wrt MSI-classification is increased (blue), while reduced in particular wrt to the project bias (orange). Fe-Bias1 dc = Squared distance correlation between learned features and project bias (orange). Fe-MSI dc = Squared distance correlation between learned features and MSI status (blue). Fe-Bias1 dc = Squared distance correlation between learned features and patient bias (green). Fe-Bias1 dc = Squared distance correlation between learned features and TMA glass bias (red).  }
    \label{fig:DC_Training}
\end{figure}

Statistical dependence, conditioned in the MSS cohort as explained in section \ref{sec:methods}, between the learned features and each bias measured with the squared distance correlation   was consistently reduced by more than half in the bias-ablated models as compared to the baseline models, as shown in Tab. \ref{table:dc fe-bias}. The $dc$ with the project bias was the most reduced one. 
During model training, as shown in Fig. \ref{fig:DC_Training}, the squared distance correlation between the learned features and the target MSI status (blue) increased as expected as a function of the number of training iterations. When comparing the two training regimes, we observed that in the baseline model (Fig. \ref{fig:DC_Training}, a)  the correlation with the MSI target increased together with all other identified biases and especially with the sample project (orange) which overlapped with the MSI target (blue). In contrast, in the model that applied the bias-ablation adversarial training (Fig. \ref{fig:DC_Training}, b), the $dc$ between the learned features and the project and patient biases did not increase, maintaining them at minimum values, while simultaneously allowing the increase of statistical dependence with the MSI target. Notably, the efficacy of the bias ablation was more marked wrt to the project bias (orange). 
\begin{table}[]
    \centering
    \caption{Statistical dependence between the learned features and each bias for the MSS group.  Quantitatively measured by the squared distance correlation ($dc$). The mean and standard deviation of the $dc$ obtained from the 5 fold baseline models is compared against the 5 fold bias-ablated models. In bold are highlighted the lowest distance correlations achieved for each bias. }
    \begin{tabular}{c | c| c| c} \toprule
    \textbf{Model} & \textbf{Project Bias} & \textbf{Patient Bias} & \textbf{Glass Bias} \\ \midrule
    Baseline model  & 0.25 $\pm0.05$
    & 0.12 $\pm 0.02$ & 0.07 $\pm 0.008$ \\
    Bias-ablation model  & \textbf{0.10} $\pm0.04$ & \textbf{0.04} $\pm0.018$ & \textbf{0.03} $\pm0.013$   \\
     \bottomrule
\end{tabular}

\label{table:dc fe-bias}
\end{table}

\begin{figure}[h]
\begin{center}

\begin{subfigure}{0.4\textwidth}
\includegraphics[width=0.9\linewidth, height=6cm]{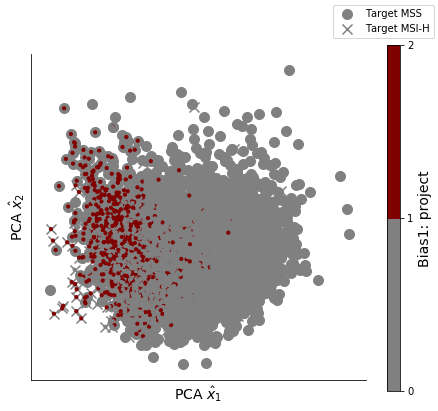} 
\caption{Baseline model features and project bias}
\label{fig:PCA_fe_project_baseline}
\end{subfigure}
\begin{subfigure}{0.4\textwidth}
\includegraphics[width=0.9\linewidth, height=6cm]{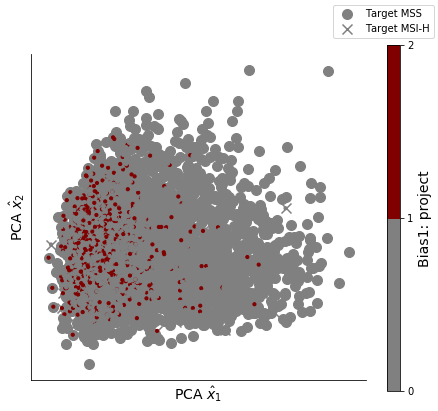}
\caption{Bias-ablated model features and project bias}
\label{fig:PCA_fe_project_bias_ablated}
\end{subfigure}
\begin{subfigure}{0.4\textwidth}
\includegraphics[width=0.9\linewidth, height=6cm]{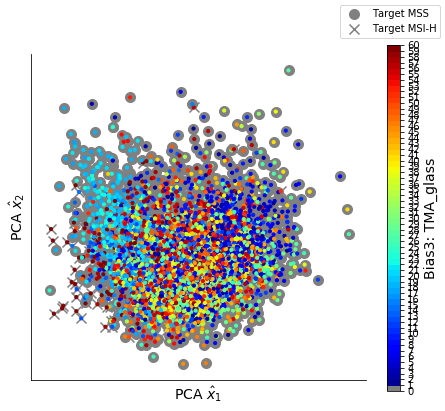} 
\caption{Baseline model features and TMA glass bias}
\label{fig:PCA_fe_TMA_baseline}
\end{subfigure}
\begin{subfigure}{0.4\textwidth}
\includegraphics[width=0.9\linewidth, height=6cm]{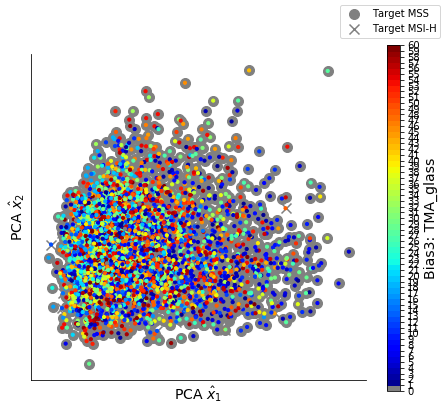}
\caption{Bias-ablated model features and TMA glass bias}
\label{fig:PCA_fe_TMA_bias_ablated}
\end{subfigure}
\caption{PCA projection of the learned features for baseline vs bias-ablated models. Color encodes the category for project bias (fig a and b) and TMA glass bias (fig c and d)}
\label{fig:PCA}
\end{center}

\end{figure}

The PCA projection of the learned representations comparing the baseline model vs the bias-ablated model is shown in Fig. \ref{fig:PCA}. The bias ablation technique results in a better representation space which is more invariant to the bias variables (project and TMA glass) while the baseline shows patterns more influenced by the bias. For example, and as illustrated in Fig. \ref{fig:PCA_fe_TMA_baseline}, most of the TMA glasses are organized as spatial clusters in the baseline model which means that the representation space of the learned features is not invariant with regards to the TMA glass. Conversely, after applying the bias-ablation technique colors of TMA glasses are less organized and follows a more homogeneous distribution \ref{fig:PCA_fe_TMA_bias_ablated}. 

\begin{figure}
    \centering
    \includegraphics[width=0.80\textwidth]{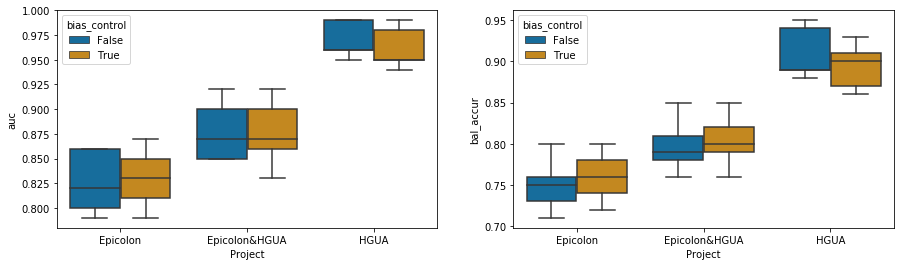}
    \caption{MSI classification results at tile level comparing the 5-fold bias-ablation models (orange) vs 5-fold baseline models (blue)  across the different projects (EPICOLON and HGUA) }
    \label{fig:boxplot_control_bias}
\end{figure}

To assess the impact of the bias-ablation technique on the MSI classification performance, we compared the validation results of 5-folds models trained for up to 3 epochs with and without the adversary ablation of the three known biases.
As shown in Fig. \ref{fig:boxplot_control_bias}, when models were evaluated on the validation set without differentiating between the projects of samples (i.e. both EPICOLON and HGUA samples included), there was not appreciable impact on performance between the bias-ablated vs baseline models at image level (both achieved an AUC of 0.87 $\pm0.03$). Nonetheless, when the baseline models were evaluated separately for each of the different projects (EPICOLON vs HGUA), the performance in HGUA samples (AUC = 0.97 $\pm0.01$, balanced accuracy = 0.91 $\pm 0.03$)  was significantly higher than in EPICOLON samples (AUC = 0.82 $\pm0.03$, balanced accuracy = 0.75 $\pm 0.03$) with a gap of up to 0.15 and 0.16 in the AUC and balanced accuracy respectively. This project gap was attenuated to 0.13 (in both the AUC and balanced accuracy) after applying the bias-ablation technique, by decreasing the performance in HGUA (AUC = 0.96 $\pm0.02$, balanced accuracy = 0.89 $\pm 0.02$) while increasing the performance in EPICOLON (AUC = 0.83 $\pm0.03$, balanced accuracy = 0.76 $\pm 0.01$). 

\subsection{Experiment 4: Results at tile vs patient level and effect of tissue types and magnifications}
For each of the 5 folds, tile predictions of the bias-controlled model were aggregated by majority voting for the decision of the MSI status at the patient level. 5-folds validation sets consisted of approximately 300 patients each and the mean prevalence of MSI-H was 9\%. In order to simulate performance on real population prevalence, metrics dependent on disease prevalence (see Table \ref{table:agg_results}) were calculated assuming a MSI-H prevalence of 15\%. 

The AUC at tile level, and including all three selected tissues (tumor epithelium, mucine and lymphocytic regions) and all 4 magnifications, was 0.87 $\pm0.03$ and increased to 0.9 $\pm0.03$ at patient level.

\begin{figure}
    \centering
    \includegraphics[width=1\textwidth]{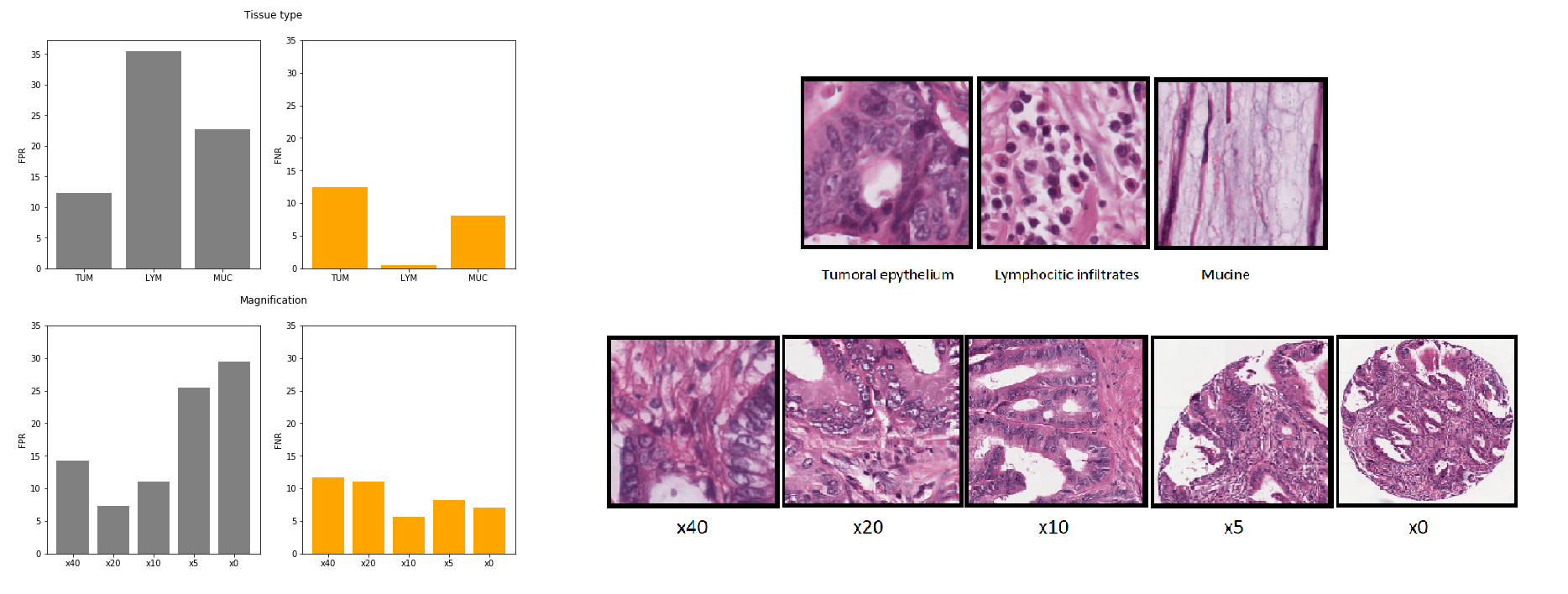}
    \caption{Average of false positive rates (blue) and false negative rates (orange) at tile level for the 5 fold bias-controlled models by tissue type (figures above) : tumor epithelium (TUM), lymphocytic infiltrates (LYM) and mucine (MUC) and different magnifications (figures below): x40, x20, x10, x5, x0}
    \label{fig:FPR_Tissue}
\end{figure}

The AUC at tile level for a bias-controlled model trained only on tumor epithelium was 0.87, but differently from the model that included all three selected tissues the AUC at patient level did not increase and remained 0.87.
In the models trained with all three tissue types, tiles tagged as lymphocytic infiltrates and mucine had an average of false positive rate (ratio of false positive over true negatives)  of 0.32 and 0.22 respectively, compared to 0.12 in tumoral epithelium tiles (see Fig. \ref{fig:FPR_Tissue}). This finding supports that nuclear and cellular characteristics of tumor epithelium are more specifics of MSI-H status, while mucine and lymphocytic infiltrates are nonspecific, which would explain the higher rates of false positives at tile level. Nonetheless, as inferred from the performance gain in AUC at patient level, regions with mucine and lymphocyte's infiltrates were probed to be relevant for MSI classification at case level by contributing to a lower rate of false negatives as shown in Fig. \ref{fig:FPR_Tissue}. Because MSI-H tumors typically exhibit larger amounts of mucine and lymphocytic infiltrates, even if not specific, it would explain the increase in model performance at patient level after applying the majority of votes.   

The AUC at tile level for bias-controlled models trained respectively at x5, x10, x20 and x40 magnifications in all three selected tissue types were 0.81, 0.85, 0.87, 0.86. According to the AUC results and as shown in Fig \ref{fig:FPR_Tissue}, the specificity was higher for x20 magnification where the lowest rates of false positives were observed. Nonetheless false negative rates decreased (lower cases missed) with decreasing magnification from x40, x20 up to x10 magnification, which argues in favor that including high level tissue architectural patterns (up to x10) contributes to increased sensitivity. The AUC at tile level for bias-controlled models and including all available magnifications was 0.87, hence the inclusion of all magnifications was non-inferior to including only one single magnification for training, and was also considered as a valuable data augmentation technique. 

\begin{table}[]
    \centering
    \caption{Patient level performance of the bias-ablated models cross-validated in 5 folds. Metrics that are dependent on MSI-H prevalence, are calculated assuming a MSI-H prevalence in the real population of 15\%. Confidence intervals for sensitivity, specificity and accuracy are "exact" Clopper-Pearson confidence intervals. Confidence intervals for the predictive values are the standard logit confidence intervals given by \citep{mercaldo2007confidence}. P = Prevalence, S = Sensitivity, E = Specificity. * Metrics that are dependent on MSI-H prevalence.}
    \begin{tabular}{c | c| c | c} \toprule
    \textbf{Metric} & \textbf{Value}  & \textbf{95\% CI} & \textbf{Definition}\\ \midrule
    AUC & 0.90 & 0.87-0.93 & \\
    Accuracy* & 88\% & 83.8\%-91.4\%& $S * P + E * (1 - P) $ \\
    Sensitivity (S)  & 87\% & 79.5\%-91.3\%& $\frac{TP}{TP+FN}$\\
    Specificity (E) & 88.3\%  & 86.5\%-89.9\%& $\frac{TN}{TN+FP}$\\
    Positive Predictive Value* & 56.5\%  & 52.6\%-60.4\%& $\frac{S*P}{S*P + (1-E)*(1-P)}$\\
    Negative Predictive Value* & 97.3\% & 96\%-98.2\%& $\frac{E*(1-P)}{E*(1-P) + (1-S)*P}$\\
    
     \bottomrule
\end{tabular}
\label{table:agg_results}
\end{table}

\subsection{Experiment 5: Visual explainability}
\begin{figure}
    \centering
    \includegraphics[width=0.80\textwidth]{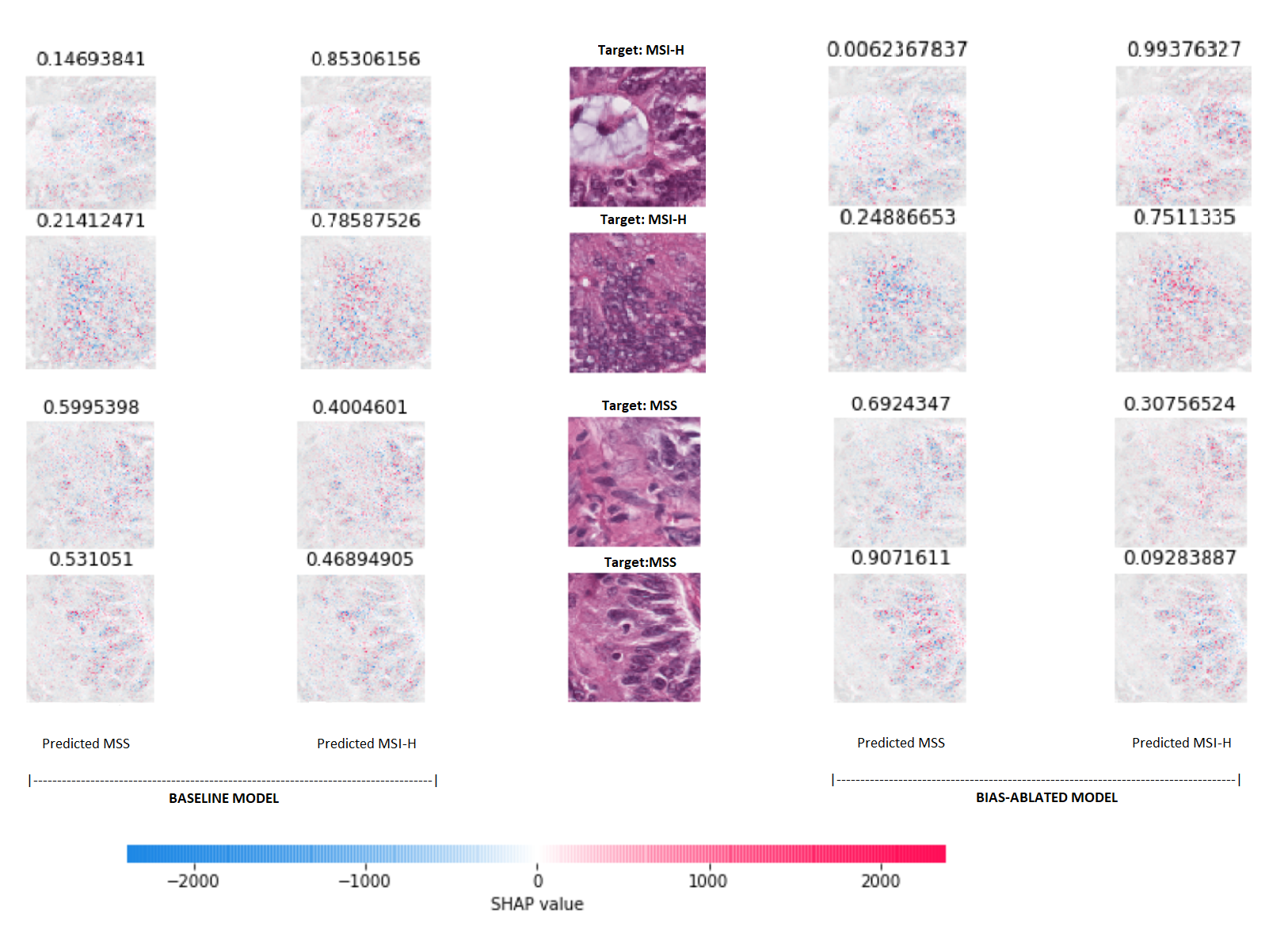}
    \caption{ SHAP (SHapley Additive exPlanations) values projected on the tiles to visually interpreting the topology and morphology of features that influence predictions of the baseline and bias-ablated model (left and right respectively). The input samples are placed in the center of the graph with their respective target labels (2 MSI-H samples at the top and 2 MSS samples below). For each of the input tiles, the resulting SHAP values obtained for both models are projected for each class on top of a grayscale copy of the input tile (left and right for MSS and MSI-H respectively). Red pixels increase the model's output while blue pixels decrease the output. The sum of the SHAP values equals the difference between the expected model output (averaged over the background dataset) and the current model output. The softmax values (prediction) for each class (MSS vs MSI-H) are shown on top of each projection.}
    \label{fig:Shap}
\end{figure}

For visually interpreting the topology and morphology of features that influenced the predictions, Fig. \ref{fig:Shap} shows that the most relevant features for classifying the samples as MSI-high vs MSS were located at conglomerates of tumoral cells with activations mainly gathered at the nucleus of the cells. Note that as a binary classification task, for the MSI-H projection images the red pixels are the most important regions influencing the prediction towards the MSI-H class, while for the MSS image projection the same pixels are shown as blue, being symmetrical. 

When comparing the projections of SHAP values between the baseline and the bias-ablation models for the same inputs, there were no perceptible differences by visual observation, nonetheless the discriminative power measured by the softmax values distance between classes is increased, in particular for MSS samples.  

\section{Discussion}
\label{sec:discussion}

We present a system for the prediction of MSI from H\&E images using artificial vision techniques that incorporates and end-to-end TMA-customized image pre-processing module to tile samples at multiple magnifications in the regions of interests guided by the automatically detected type of tissues and a multiple bias distiller system integrated with the MSI predictor. 

In the present work we find that TMAs have special characteristics, not reported for WSIs, which make them specially challenging for the application of DL methods in digital pathology, emphasizing the relevance of addressing biases. 

A systematic study of biases at tile level demonstrated three hidden variables interfering with the model's learned representations: the project of origin of samples, the patient's spot and the TMA glass where each spot was placed. Even if it is preferred to control for any of those types of biases at the dataset level, it entails either obtaining more tissue and/or re-allocating them in new TMA glasses which was unfeasible in general. Instead, we reused the TMAs as they were provided for research purposes given that first, the most optimal management of tissue samples avoiding sample waste is always desirable and second, the presence of associations spurious or otherwise undesirable that are exploited by DL models, rather than being an odd problem affecting only our work, is a general, common and not yet resolved challenge in medical datasets used to train AI systems. Consequently, we decided to dedicate the efforts to systematically study and address the biases at the learning stage. For this purpose, a novel multiple bias rejecting technique has been implemented at the deep learning architecture to directly avoid learning the batch effects introduced by protected variables.

The implemented method achieved a significant reduction in the dependence of the learned features with regards to the project bias and patient bias in the general study populations but did not reduced the glass bias dependence in the HGUA project where we found that for most of the TMAs there were only samples with one single type of target class included. We observed that the ablation method is highly effective for mitigating bias in datasets where for all possible ordered pairs of protected variable classes and target classes there are representative samples even if heavily in-balanced. In addition, the performance in the MSI classification improved in terms of AUC and balanced accuracy for the population meeting this condition (EPICOLON cohorts).  Conversely,  when this condition was not meet (as in the case of the HGUA for the glass bias, where the target and protected variable had an unequivocal association in the samples), the statistical dependence was not eliminated and the classification performance in the target task still seemed to exploit the bias maintaining a highly marked predictive advantage in comparison with the classification results in the EPICOLON cohorts. When analyzed considering all study population, the learned features from the bias-ablated model had maximum discriminative power with respect to the task and minimal statistical mean dependence with the biases.

In contrast to other population-based cohorts  where MSI prevalence is around 15\%, in the EPICOLON project, the cohorts had a mismatch repair deficiency in only 7.4\% patients. In EPICOLON I only 91 patients (7.4\%) had a mismatch repair deficiency with tumors exhibiting either genetic microsatellite instability (n=83) or loss of protein expression (n=81) \citep{pinol2005accuracy}. The underlying reason for this difference, which was only observed in the non-hereditary MSI subset, is not yet studied but epidemiological reasons such as the effect of Mediterranean diet can not be excluded as a cause of half the prevalence of non-hereditary MSI frequency as compared to other regions.  

Regarding the classification MSI status at case level, we observed that the performance consistently increased in all experiments when not only tumor epithelium but also the mucinous component and lymphocytic infiltrate regions were included. Those regions were probed to be non-specific at image level, but increased the sensibility at patient level, which is a desirable characteristic for screening purposes.

Also, a x20 magnification achieved the higher specificity, but at the same time, reducing magnifications up to x10 contributed to higher sensitivity of the models. This observation supports that the best approach would be to include different magnifications, helping the model to learn both low and high level tissue architectural patterns at the same time. Moreover, including all magnifications during training was considered in this project as a data augmentation technique. 

The tissue classifier module reached an AUC of 0.98 in the validation set. This module was capable of classifying the different regions of the image: tumor epithelium, stroma, normal epithelium, mucine, muscular fibers, lymphocytic infiltrates, debris, adipose tissue and background. Regarding the final task of MSI status prediction, the AUC at tile level, including all three selected tissues (tumor epithelium, mucine and lymphocytic regions) and all magnifications, was 0.87±0.03 and increased to 0.9±0.03 at patient level.

Limitations of the study are as follow:
We observed, after applying the image pre-processing pipeline, a reduction in the final available MSI population, altering its original frequency in the EPICOLON project. Specifically, up to 9\% ( n = 159 patients) in the EPICOLON population were excluded from the final analysis, where the largest proportion of excluded patients corresponded to the MSI-H arm, hence reducing its frequency from the expected 7.4\% in EPICOLON to 5.8\% (n = 105). This was explained by a more intensive molecular tissue testing performed in MSI-H cases in the context of the EPICOLON research project that exhausted a subset of spots with no viable regions of epithelial tumor left. The reduction in the final available MSI population, altering its original frequency, was consequently not longer considered a representative sample of the original EPICOLON MSI-H population. To overcome this limitation, the dataset was further enriched with cases from a single hospital as shown in Fig \ref{fig:Sample}.  On the one hand, this enrichment would increase the exposure of the MSI classifier to additional unselected MSI-H cases and, on the other hand, the bias rejecting technique implemented successfully addressed the batch effect introduced by the project of origin. Finally, result metrics which are impacted by disease prevalence, are calculated considering the MSI-H prevalence in the real population (15\%), so as to approximate its performance in the clinical setting.

As future work, only after addressing the remaining TMA-glass bias at the data level, testing the generalizability of the system in an independent and prospective test would be necessary. Also multimodal variables including age, stage, location, Beth criteria could be included in the model to explore their potential to improve the predictive capacity.

\section*{Acknowledgement}

This work is supported by the Programa Estatal de Generación de Conocimiento y Fortalecimiento del Sistema Español de I+D+i, financed by the Instituto de Salud Carlos III and Fondo Europeo de Desarrollo Regional 2014-2020 (Grant DTS19/00178).

We want to particularly thank the patients and the BioBank ISABIAL integrated in the Spanish National Biobanks Network and in the Valencian Biobanking Network for their collaboration. 




\ifdefined\arxiv
    \bibliographystyle{unsrt}
\else
    \bibliographystyle{model2-names}\biboptions{authoryear}
\fi

\newpage
\bibliography{main.bib}

\clearpage
\input{appendix.tex}










\end{document}

%% file: appendix.tex
\appendix

\bigbreak